\documentclass[10pt]{article} 
\usepackage[accepted]{rlc}

\usepackage{amssymb}            
\usepackage{mathtools}          
\usepackage{mathrsfs}           
\usepackage{graphicx}           
\usepackage{subcaption}         
\usepackage[space]{grffile}     
\usepackage{url}                

\usepackage{paralist}
\newcommand{\contG}{\tilde{G}}
\newcommand{\dt}{\Delta}
\newcommand{\defeq}{\overset{\textrm{def}}{=}}

\title{An Idiosyncrasy of Time-discretization in\\Reinforcement Learning}

\author{Kris De Asis  \\
    kldeasis@ualberta.ca \\
    Department of Computing Science\\
    University of Alberta
    \And
    Richard S. Sutton \\
    rsutton@ualberta.ca\\
    Department of Computing Science\\
    University of Alberta}

\begin{document}

\maketitle

\begin{abstract}
Many reinforcement learning algorithms are built on an assumption that an agent interacts with an environment over fixed-duration, discrete time steps. However, physical systems are continuous in time, requiring a choice of time-discretization granularity when digitally controlling them. Furthermore, such systems do not wait for decisions to be made before advancing the environment state, necessitating the study of how the choice of discretization may affect a reinforcement learning algorithm. In this work, we consider the relationship between the definitions of the continuous-time and discrete-time returns. Specifically, we acknowledge an idiosyncrasy with naively applying a discrete-time algorithm to a discretized continuous-time environment, and note how a simple modification can better align the return definitions. This observation is of practical consideration when dealing with environments where time-discretization granularity is a choice, or situations where such granularity is inherently stochastic.
\end{abstract}

\section{Introduction}

Reinforcement learning provides a framework for solving sequential decision making problems based on evaluative feedback \citep{sutton2018rlbook}. It remains a promising approach for robot learning as it can allow for real-time adaptation of behavior. Many reinforcement learning algorithms assume that the agent-environment interaction occurs at synchronous, discrete time steps, where the environment waits for an action before advancing. In contrast, real-world physical systems are continuous in time, and do not wait for an agent's input. As such, time-discretization becomes a necessary and important consideration, as evidenced by \cite{mahmood2018settingup}.

Prior work suggests that current reinforcement learning algorithms are sensitive to the choice of discretization. \cite{tallec2019} emphasize that action-values converge to state-values as the discretization interval approaches zero, creating degenerate cases for algorithms like Q-learning. Similarly, \cite{munos2006contpg} showed that the variance of policy gradients can be infinite under the same limit. \cite{zhang2023resolution} characterize a fundamental bias-variance trade-off with the degree of discretization while \cite{mahmood2018settingup} detail another trade-off between having fine-grained control and being able to discern the changes between subsequent states.  Finally, \cite{farrahi2023reducing} provide guidelines for time-discretization-aware parameter selection by acknowledging how changes in discrete-time parameters influence the underlying continuous-time objective.

In this work, we explicitly view the discrete-time objective as a discrete approximation of the continuous-time objective. By considering \textit{when} rewards occur, particularly in existing continuous-control environment setups, we identify an idiosyncratic dependence on the choice of discretization beyond those listed by \cite{tallec2019} and \cite{farrahi2023reducing}. Specifically, the discrete-time return can be viewed as mixing two Riemann sums. We characterize and demonstrate that this is a relatively poor integral approximation in comparison with a conventional Riemann sum and provide a simple modification to the definition of the return to better align the objectives.

The contributions of this work are as follows:
\begin{compactitem}
    \item We acknowledge and characterize an issue with naively applying a discrete-time reinforcement learning algorithm to a \textit{discretized} continuous-time environment in terms of a discrepancy between the discrete-time and continuous-time definitions of the return.
    \item Based on an integral approximation perspective, we propose a simple modification to the definition of the return to alleviate this idiosyncratic dependence on time-discretization.
    \item We characterize when the modification will have a modest impact and support our claim with empirical evaluation in both continuous-time prediction and control.
\end{compactitem}

\section{Definitions of the Return}

In discrete-time reinforcement learning, the discounted return from time step $t$ onward is defined as:
\begin{equation}
    G_t = \sum_{k=t}^{T - 1}{\gamma^{k - t} R_{k + 1}},
    \label{eqn:discretereturn}
\end{equation}
where $T$ is the final time step of an episodic task, or $\infty$ in an infinite-horizon setting. In continuous-time reinforcement learning (e.g., \citealp{doya2000contrl,mehta2009pontryagin,fremaux2013contac,lee2021contpi,tallec2019}), we instead define the \textit{integral return} from time step $t$ onward:
\begin{equation}
    \contG_t = \int_t^T{\gamma^{\tau-t}R_\tau d\tau}.
    \label{eqn:integralreturn}
\end{equation}
This formulation is pertinent to applications with real-time interaction (e.g., robotics). Despite being continuous in time, robots are often digitally controlled, necessitating understanding the impact of the choice of time-discretization and how it relates these two objectives.

\section{When Rewards Occur}

There are notational differences in the literature with respect to time indices in the discrete time return (Equation \ref{eqn:discretereturn}). Some define it to start from $R_{t+1}$ (e.g., \citealp{sutton1988td,precup2000pdis,vanseijen2009expsarsa,barreto2017successor}), as presented in this document, while some would start from $R_{t}$ (e.g., \citealp{watkins1989qlearning,vanhasselt2010doubleq,mnih2015dqn,wang2016dueling}). This inconsistency is inconsequential when solely considering the discrete-time setting as the rewards occur at the same locations in an agent's stream of experience. However, it has implications when viewed as a discrete approximation to an underlying integral return. Thus, it is worth considering \textit{when} rewards occur.

We emphasize the focus on a setting where there is an underlying continuous-time objective of which a digital learning agent samples at an arbitrary (and potentially variable) frequency. Despite the discrete-time notational differences, it is often agreed upon that from the agent's perspective, the reward and next state are jointly observed. This is reflected in environment step calls in relatively standard reinforcement learning APIs (e.g., \citealp{brockman2016gym}), agent-environment interaction diagrams (e.g., \citealp{sutton2018rlbook}), or explicit acknowledgement that reward can be a function of state, action, and \textit{next state} (e.g., \citealp{puterman1994mdps}). In real-time settings that do not wait for an agent's input, meaningful evaluative feedback must come \textit{after} time $t$ as actions take time to execute and to have a causal influence. Hardware limitations on sampling rates further delay when a system can receive feedback for an action. In many existing robotics environments, where the considered setting is especially pertinent, rewards are often explicitly computed based on the next time step's state information. For example, rewards based on distance traveled in some direction between two time steps, or distance between an end-effector and a desired setpoint at the subsequent time step, as done by \cite{todorov2012mujoco}, \cite{brockman2016gym}, and \cite{mahmood2018benchmarking}.

Of note, semi-MDPs and options \citep{sutton1999options,precup2000options} address the problem of when rewards occur, but under the assumption that one has access to higher-frequency interaction with the environment to integrate the discounted sum of rewards within the discretization interval. It is akin to the agent being aware of and able to time when each component of a temporally-extended reward occurs. Here, we consider when one \textit{does not} have access to these higher-frequency samples but is aware of how much time has elapsed between discrete decision points. Acquiring such information may not be possible due to hardware limitations, and highlights a nuance that arises when naively applying a discrete-time algorithm to a discretized continuous-time environment.

\section{Implications for Time Discretization}

\begin{figure}[ht]
  \centering
  \begin{subfigure}[b]{0.32\textwidth}
    \centering
    \includegraphics[width=\textwidth]{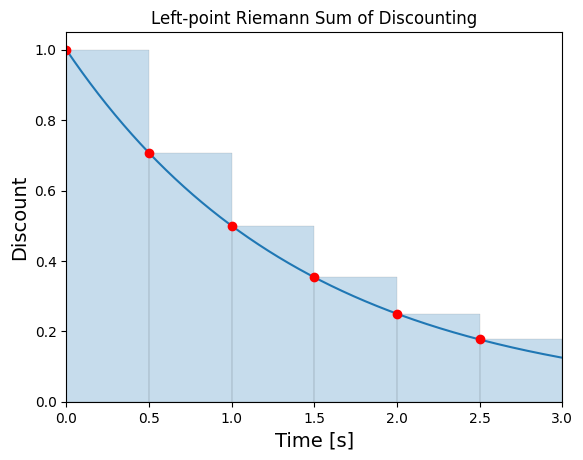}
  \end{subfigure}
  \begin{subfigure}[b]{0.32\textwidth}
    \centering
    \includegraphics[width=\textwidth]{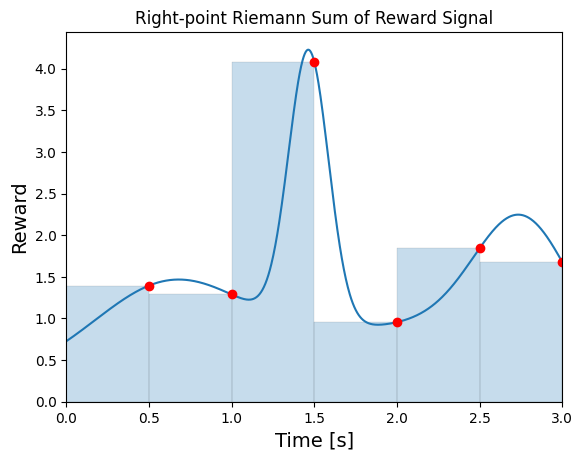}
  \end{subfigure}
  \begin{subfigure}[b]{0.32\textwidth}
    \centering
    \includegraphics[width=\textwidth]{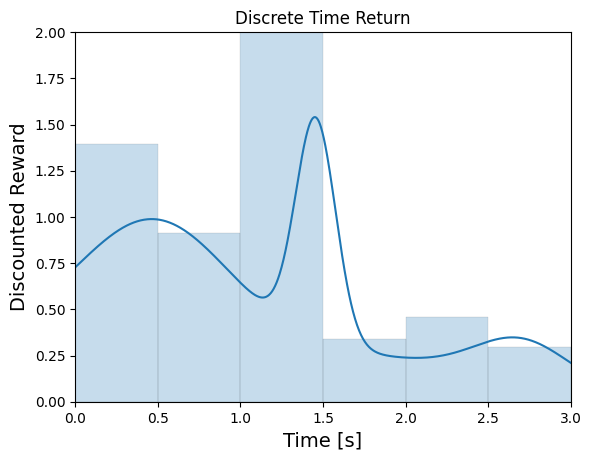}
  \end{subfigure}
  \caption{The resulting sum when applying a discrete-time algorithm to a discretized continuous-time domain. Note how rectangle heights may fall out of the function's range within an interval.}
  \label{fig:lprpexample}
\end{figure}

\begin{figure}[ht]
  \centering
  \includegraphics[width=0.45\textwidth]{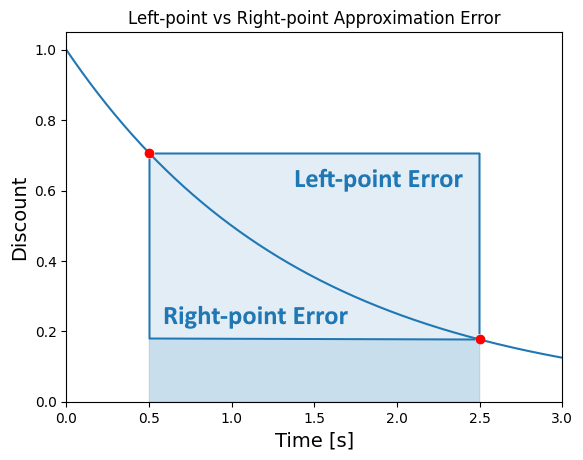}
  \caption{A visualization of the left-point and right-point Riemann sum approximation errors for an exponential decay. Due to curvature, a right-point Riemann sum will always have lower error.}
  \label{fig:lprperr}
\end{figure}

If we consider rewards jointly arriving with the next state, at least from the agent's perspective, then there is an idiosyncrasy with respect to approximating an underlying integral return. While definitions of the discrete-time return may differ in their use of reward time indices, they are consistent on when discounting begins: the first reward is given weight $\gamma^0 = 1$, with subsequent rewards weighted by increasing powers of $\gamma$. We can view the integral return in Equation \ref{eqn:integralreturn} to be of the form:
\begin{equation}
    \int_{t}^{T}{f(\tau)g(\tau)d\tau},
    \label{eqn:integralproduct}
\end{equation}
where $f(\tau)$ is the discounting term and $g(\tau)$ is the reward signal. A right-point Riemann sum approximation to this would yield:
\begin{equation}
    \sum_{i=0}^{n - 1}{f(\tau_i)g(\tau_i)\dt},
    \label{eqn:rightriemann}
\end{equation}
where $\dt = \frac{T - t}{n}$ and $\tau = \{t + \dt, t + 2\dt, ..., T\}$. The right-point Riemann sum beginning with $t + \dt$ aligns with an agent jointly receiving a reward with the observation of the next state. However, this sum would weight the first reward by $\gamma^{\dt} \neq \gamma^0$. This highlights that if one naively applies a discrete-time reinforcement learning algorithm to a discretized continuous-time environment, it is akin to a left-point Riemann sum for discounting and a right-point Riemann sum for rewards:
\begin{equation}
    \sum_{i=0}^{n - 1}{f(\tau_i)g(\tau_{i + 1})\dt},
    \label{eqn:weirdmann}
\end{equation}
where $\tau \in \{t, t + \dt, t + 2\dt, ... , T\}$. See Figure \ref{fig:lprpexample} for a visualization of this Riemann sum. This sum still converges to the correct integral as $n\rightarrow\infty$ as Bliss's Theorem \citeyearpar{bliss1914integration} shows that each function may be evaluated \textit{anywhere} in the interval. However, for the specific case where a left-point Riemann sum is used for discounting, we expect this to perform \textit{worse} than committing to a right-point Riemann sum. If one draws a rectangle with opposite corners at any two points of an exponential decay, the area above and below the curve represents the approximation errors of left- and right-point Riemann sums, respectively. There will always be more area above the curve than below due to the curvature of exponential decay, implying that an underestimate (right-point) has strictly lower error than an overestimate (left-point). This is visualized in Figure \ref{fig:lprperr}.

To rectify this discrepancy and commit to a right-point Riemann sum approximation, we simply multiply the discrete-time return by a factor of $\gamma$. For example, with $\dt = 1$:
\begin{equation}
    \gamma G_t = \gamma R_{t+1} + \gamma^2 R_{t+2} + \cdots.
\end{equation}
For a \textit{fixed, pre-specified} action cycle-time $\dt$, there is no loss of generality, as the discrete-time return is proportional by a factor of $\gamma^{\dt} \dt$. However, this is not the case when $\Delta$ may vary over time, for example, due to an adaptive algorithm (e.g., \citealp{karimi2023dynamic}) or inherent stochasticity. These concerns similarly apply to a variable $\gamma$ and may extend toward tuning fixed-$\Delta$ or $\gamma$ in practice in terms of an unintuitive dependence on discretization. To emphasize the dependence on $\dt$, we note the more explicit definition of the right-point Riemann sum return:
\begin{align}
    G_t^{RP} &\defeq \sum_{k=t}^{T-1}\gamma^{\sum_{i=t}^{k}{\dt_{i+1}}}R_{k+1}\dt_{k+1} \label{eqn:rpreturn} \\
    &= \gamma^{\dt_{t+1}} R_{t+1} \dt_{t+1} + \gamma^{\dt_{t+1} + \dt_{t+2}} R_{t+2} \dt_{t+2} + \cdots . \nonumber
\end{align}

\cite{tallec2019} and \cite{farrahi2023reducing} have acknowledged the modifications of scaling rewards by $\Delta$ and exponentiating $\gamma$ by $\Delta$ in terms of improving robustness to time-discretization. The key difference and contribution in Equation \ref{eqn:rpreturn} being the earlier discounting.

\section{Comparison with Standard Riemann Sums}
\label{sec:sumcomparisons}

To see how the discrete-time return (DTR) in Equation \ref{eqn:weirdmann} compares with a right-point Riemann sum, we use them to numerically integrate random continuous-time signals. Inspired by robotics, we consider periodic signals and Gaussian mixtures. Periodic signals are comparable to signals pertaining to robot locomotion, while Gaussian mixtures instead resemble both sparse and distance-based rewards depending on the spread of each Gaussian. We fix the signal length to 3 seconds, with no loss of generality due to being continuous in time. Each signal generator is detailed below:

\textbf{Random Periodic Signals} - This signal sums 6 sinusoids $\sum_{i=0}^{5}{A_i \sin (\omega_i t + \phi_i)}$ with angular frequencies $\omega \in \{\frac{2\pi}{4}, \frac{2\pi}{2}, 2\pi, 4\pi, 8\pi, 16\pi\}$, amplitudes $A_i \sim \mathcal{N}(0, 1)$, and phase shifts $\phi_i \sim \mathcal{U}(0, 2\pi)$.

\textbf{Random Gaussian Mixtures} - This signal sums 6 Gaussians $\sum_{i=0}^5{\mathcal{N}(\mu_i, \sigma_i)}$ with means $\mu_i \sim \mathcal{U}(0, 3)$ and standard deviations $\sigma_i \sim \mathcal{U}(0, \frac{3}{2})$.

\begin{figure}[ht]
  \centering
  \begin{subfigure}[b]{0.32\textwidth}
    \centering
    \includegraphics[width=\textwidth]{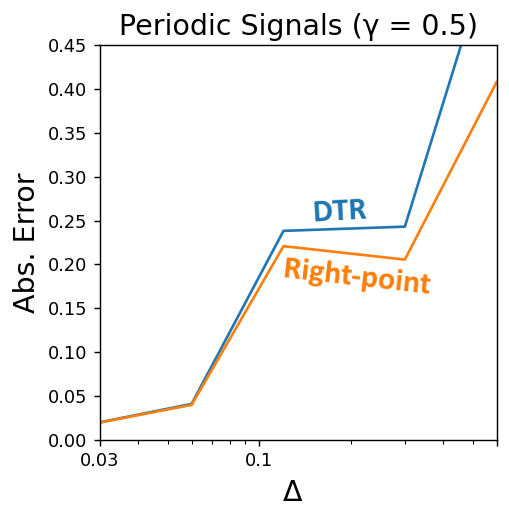}
  \end{subfigure}
  \begin{subfigure}[b]{0.32\textwidth}
    \centering
    \includegraphics[width=\textwidth]{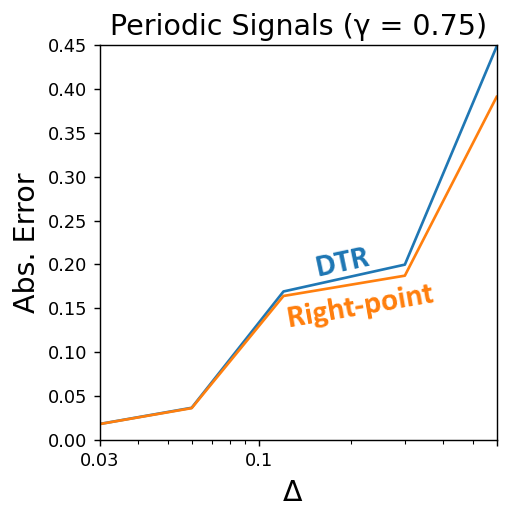}
  \end{subfigure}
  \begin{subfigure}[b]{0.32\textwidth}
    \centering
    \includegraphics[width=\textwidth]{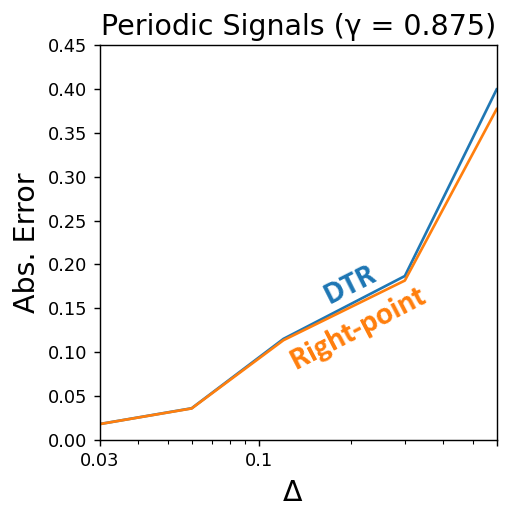}
  \end{subfigure} \\
  \begin{subfigure}[b]{0.32\textwidth}
    \centering
    \includegraphics[width=\textwidth]{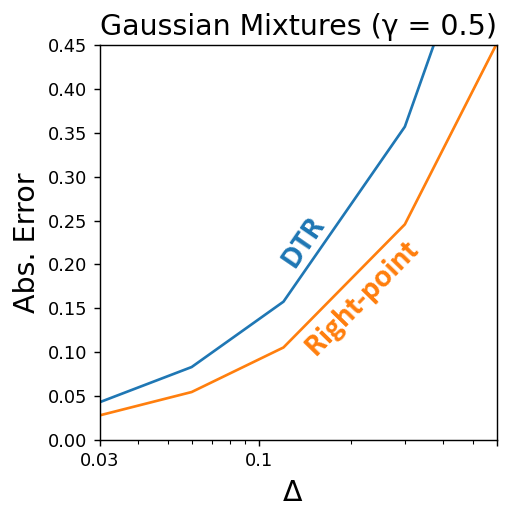}
  \end{subfigure}
  \begin{subfigure}[b]{0.32\textwidth}
    \centering
    \includegraphics[width=\textwidth]{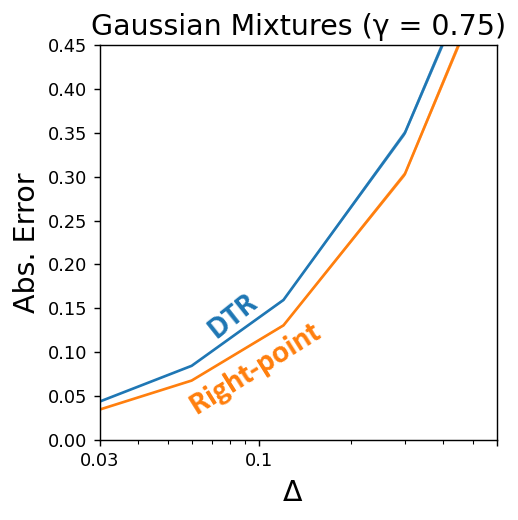}
  \end{subfigure}
  \begin{subfigure}[b]{0.32\textwidth}
    \centering
    \includegraphics[width=\textwidth]{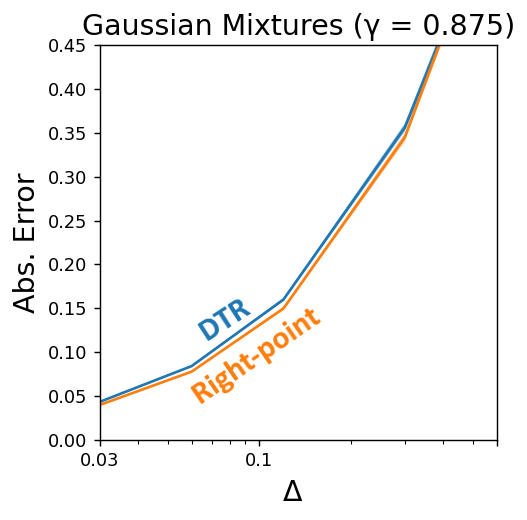}
  \end{subfigure}
  \caption{Numerical integration approximation error on \textit{discounted} random signals. Results are averaged over $10^6$ signals and shaded regions represent one standard error.}
  \label{fig:exp_riemanncomparisons}
\end{figure}

For each method, we varied the number of intervals $n \in \{5, 10, 25, 50, 100\}$, the discount factor $\gamma \in \{0.5, 0.75, 0.875\}$, and measured the absolute error of the integral approximation relative to a fine-grained mid-point Riemann sum with $10^4$ intervals. The values of $\gamma$ used may appear small and unrepresentative of typical values. We however note that the discount is \textit{per second} and that for a robot sampling every 30 ms, $\gamma=0.5$ is effectively $\gamma^\dt=0.5^{0.03}\approx0.98$ per discrete time step. Averaged across $10^6$ randomly generated signals of each type, the results can be seen in Figure \ref{fig:exp_riemanncomparisons}.

As expected, the errors generally increase as $\Delta \propto \frac{1}{n}$ increases. There is a consistent dip in error with the periodic signals which is likely due to the intervals coincidentally aligning with the pre-specified frequencies. Across all settings, DTR had larger absolute error and is consistent with our hypothesis that DTR would perform worse than right-point when integrating discounted signals. The gap closes as $\gamma \rightarrow 1$ as the sums are equivalent at this extreme.

\begin{figure}[ht]
  \centering
  \begin{subfigure}[b]{0.32\textwidth}
    \centering
    \includegraphics[width=\textwidth]{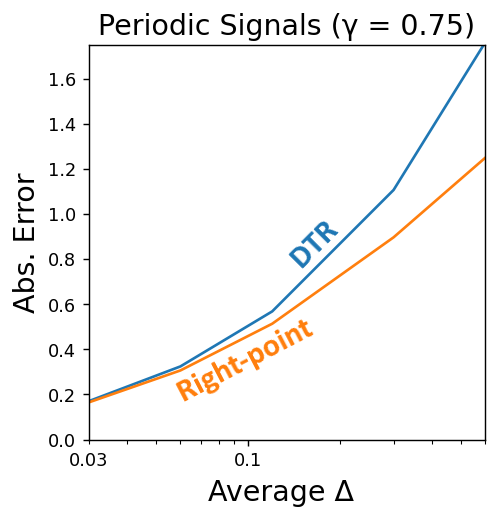}
  \end{subfigure}
  \begin{subfigure}[b]{0.32\textwidth}
    \centering
    \includegraphics[width=\textwidth]{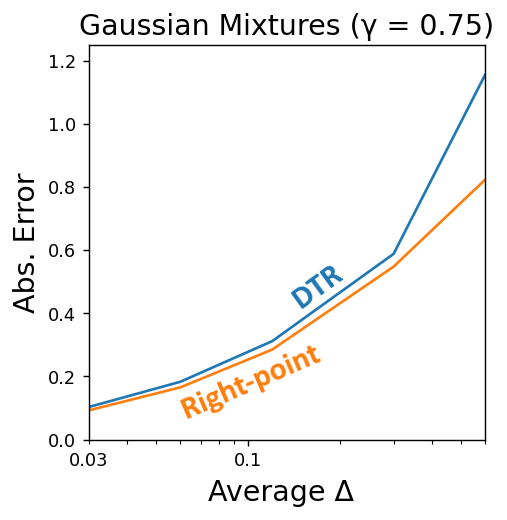}
  \end{subfigure}
  \caption{Numerical integration approximation error on \textit{discounted} random signals, with \textit{stochastic discretization intervals}. Results are averaged over $10^6$ signals and shaded regions represent one standard error.}
  \label{fig:exp_stochasticintervals}
\end{figure}

We then considered stochastic intervals to simulate variable time-discretization. This was implemented by sampling, sorting, and re-scaling a set of $n + 1$ uniform random points to represent interval endpoints. This is particularly pertinent as DTR is no longer proportional to right-point and reflects the variability in applications on real-time systems. Fixing $\gamma = 0.75$, Figure \ref{fig:exp_stochasticintervals} shows results averaged across $10^6$ randomly generated signals of each type plotted against \textit{average} $\Delta$. Errors generally increased, with DTR maintaining larger approximation error across every setting.

\begin{figure}[ht]
  \centering
  \begin{subfigure}[b]{0.32\textwidth}
    \centering
    \includegraphics[width=\textwidth]{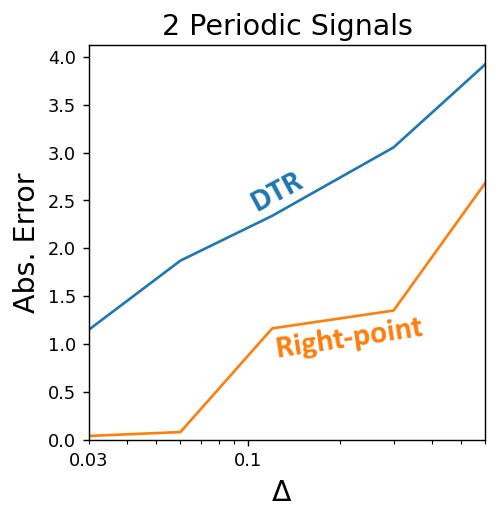}
  \end{subfigure}
  \begin{subfigure}[b]{0.32\textwidth}
    \centering
    \includegraphics[width=\textwidth]{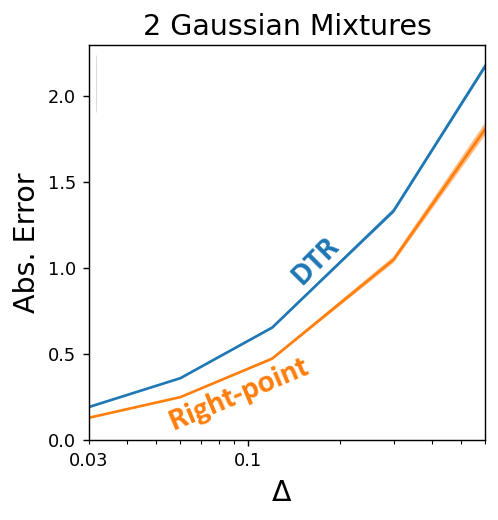}
  \end{subfigure}
  \begin{subfigure}[b]{0.32\textwidth}
    \centering
    \includegraphics[width=\textwidth]{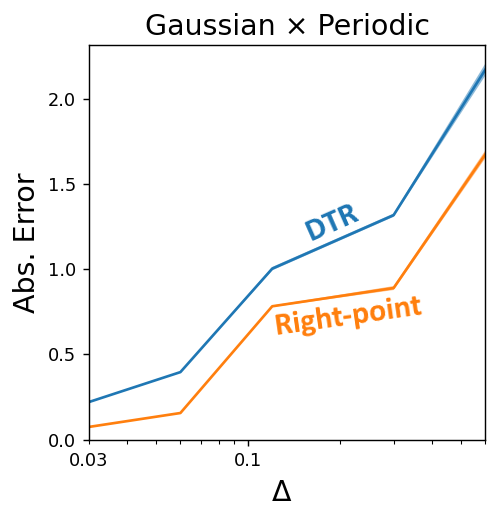}
  \end{subfigure}
  \caption{Numerical integration approximation error on \textit{undiscounted} products of random signals. Results are averaged over $10^6$ signals and shaded regions represent one standard error.}
  \label{fig:exp_nondiscounting}
\end{figure}

Lastly, to see whether results hold beyond exponential discounting, we considered the product of each pair of the signal generators. This evaluates each sum in a more general numerical integration setting, while resembling transition-dependent $\gamma$ as \cite{white2016taskspec} has advocated for in reinforcement learning. Averaged across $10^6$ randomly generated signal pairs, the results can be seen in Figure \ref{fig:exp_nondiscounting}. Perhaps surprisingly, the gap between DTR and the right-point Riemann sum widens dramatically. This suggests that beyond the structure of discounting, DTR is a generally worse integral approximation.

A key takeaway from these results is that shifting the discount factor in the discrete-time return yields a better prediction target (e.g., in value-based methods) in terms of error between the integral return. To reiterate, in the \textit{fixed $\Delta$} case, the sums are proportional despite the gaps in approximation error. This suggests that the improvement is inconsequential for control. However, in the \textit{variable $\Delta$} setting, we expect that learning from estimates which better approximate the underlying integral return should improve the capability to maximize it. We explore this further in the next section.

\section{Discretized Continuous-time Control}

To evaluate the right-point Riemann sum in a continuous-time control setting, we build off of the REINFORCE algorithm \citep{williams1992reinforce}. Such a choice is due to the algorithm's simplicity, allowing for more confidence in attributing differences in performance. We specifically use \textit{online} REINFORCE with eligiblity traces \citep{kimura1995ipg} and dropped the $\gamma^t$ term:
\begin{align*}
    \mathbf{z} &\leftarrow \mathbf{z} + \nabla_\theta \log{\pi(A_t|S_t)} \\
    \theta &\leftarrow \theta + \alpha R_{eff} \mathbf{z} \\
    \mathbf{z} &\leftarrow \gamma^{\Delta_{t+1}} \mathbf{z} ,
\end{align*}
where $\Delta_{t+1}$ is the elapsed time between time steps $t$ and $t+1$, $R_{eff} = R_{t+1} \Delta_{t+1}$ for the discrete-time return, and $R_{eff} = \gamma^{\Delta_{t+1}} R_{t+1} \Delta_{t+1}$ for the right-point Riemann sum. The above algorithm employs the recommendations of \cite{farrahi2023reducing} for making algorithms more robust to time-discretization, emphasizing that the proposed right-point modification is complimentary. 

We designed a simulated \textit{Servo Reacher} environment based on the setup by \cite{mahmood2018benchmarking}, with physical parameters sourced from a Dynamixel MX-28AT data sheet. This custom environment allows for fine-grained computation of the integral return, and flexibility in the discretization intervals an agent can sample at. Full environment specification can be found in Appendix \ref{sec:app_env_details}. To simulate the inherent stochasticity of a real robot, Gaussian noise was added to the target discretization interval, $\Delta_t \sim \mathcal{N}(\Delta_\mu,10 \textrm{ ms})$, with a hard minimum interval of 1 ms. We additionally included a 1\% chance to sample the interval from $\mathcal{N}(1000 \textrm{ ms},10 \textrm{ ms})$ to simulate ``catastrophic'' events akin to communication errors. Of note, in less-exhaustive experiments not presented, such catastrophic events did not strongly impact or change the conclusions of the results.

\begin{figure}[ht]
  \centering
  \begin{subfigure}[b]{0.4\textwidth}
    \centering
    \includegraphics[width=\textwidth]{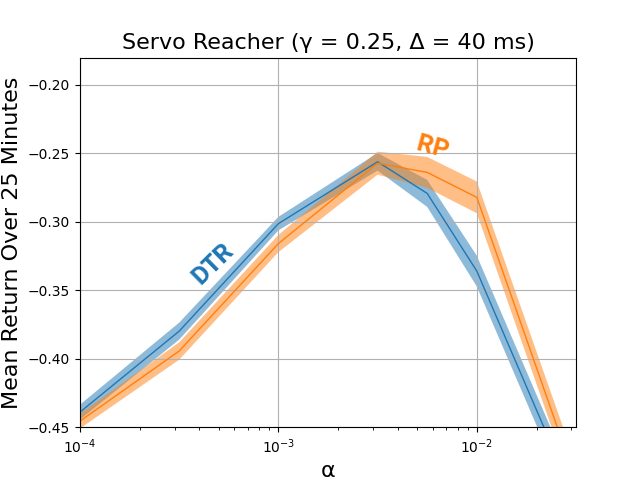}
    \caption{Parameter Sensitivity ($\Delta_\mu = 40 \textrm{ ms}$)}
  \end{subfigure}
  \begin{subfigure}[b]{0.4\textwidth}
    \centering
    \includegraphics[width=\textwidth]{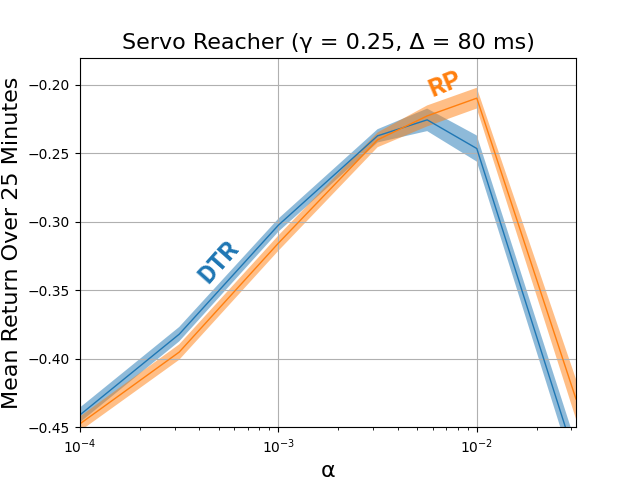}
    \caption{Parameter Sensitivity ($\Delta_\mu = 80 \textrm{ ms}$)}
  \end{subfigure}
  \begin{subfigure}[b]{0.4\textwidth}
    \centering
    \includegraphics[width=\textwidth]{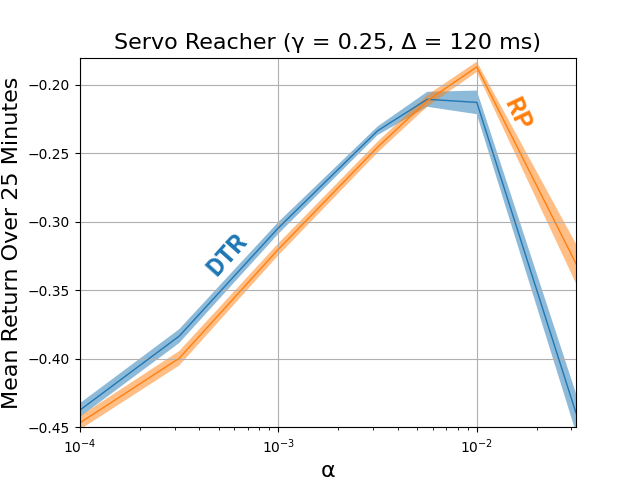}
    \caption{Parameter Sensitivity ($\Delta_\mu = 120 \textrm{ ms}$)}
  \end{subfigure}
  \begin{subfigure}[b]{0.4\textwidth}
    \centering
    \includegraphics[width=\textwidth]{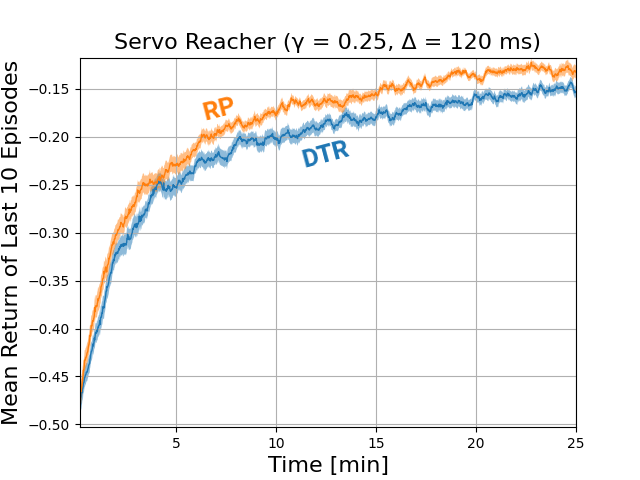}
    \caption{Learning Curves ($\Delta_\mu = 120 \textrm{ ms}$)}
  \end{subfigure}
  \caption{Servo Reacher results for REINFORCE using the discrete-time return (DTR) and right-point Riemann sum (RP), averaged over 100 runs. Shaded regions represent one standard error.}
  \label{fig:exp_ipg}
\end{figure}

Each agent's policy used a two-hidden-layer fully-connected network with $tanh$ activations, with its output being treated as the mean of a Gaussian with an initial (bias unit) standard deviation of 1. We fixed $\gamma = 0.25$, which when using an interval of 40 ms, corresponds with $\gamma^{0.04} \approx 0.95$ per discrete time step. We considered target discretization intervals $\Delta_\mu \in \{40, 80, 120\}$ ms with a 4 second time limit and measured the episodic integral return. Averaged over 100 runs of 25 (simulation) minutes, Figure \ref{fig:exp_ipg} shows parameter sensitivity curves and the best parameters' learning curves.

An initial observation is a systematic lag between the sensitivity curves of the two algorithms at low $\alpha$. This is due to the return magnitudes being roughly proportional by a factor of $\mathbb{E}[\gamma^{\Delta_t}]$. If one absorbs this factor into the step-size, the right-point Riemann sum can be viewed as using a smaller \textit{effective} $\alpha$ in the policy gradient update. Scaling the figure to use this effective $\alpha$ can be found to align the curves at low $\alpha$. Nevertheless, we find that after accounting for this shift, REINFORCE with the right-point Riemann sum never performed worse and can significantly outperform the discrete-time return with both algorithms properly tuned. The right-point Riemann sum is seen to improve with \textit{increasing} $\Delta_\mu$, in line with the approximation error results in Section \ref{sec:sumcomparisons}. Acknowledging that the two returns are roughly proportional by $\mathbb{E}[\gamma^{\Delta_t}]$, the results support that improvements are expected as this term deviates from 1 (i.e., decreasing $\gamma$ or increasing $\Delta_\mu$).

\section{Conclusions and Future Work}

In this work, we identified and characterized an idiosyncrasy of time-discretization in reinforcement learning. Specifically, a nuance between the definitions of the discrete-time and continuous-time returns when viewing one as a discretization of the other. Our results suggest that when one does not have access to evaluating the integral return via options, one can better align the objectives by shifting the discount factor to begin discounting sooner. This provides \textit{unification} in that the discrete-time return becomes a relatively straight-forward discretization of the integral return. We strongly emphasize the \textit{simplicity} of the modification and how apart from the $\gamma = 0$ extreme, such a modification has no loss of generality in discrete-time or with fixed discretization intervals due to proportionality with the conventional discrete-time return. The returns are equivalent as $\gamma^{\Delta}\rightarrow 1$, but as $\gamma^{\Delta}$ deviates from 1, the right-point return is a better prediction target in terms of integral approximation error and improves control performance with \textit{variable} time-discretization. Beyond integral approximation, the modification has intuitive appeal in that results from catastrophically long delays are attenuated in the return, rather than fully crediting an action for that outcome.

This work assumed that rewards better align with the subsequent time-step, as is often the case in the setups of existing continuous-time environments. However, should there be domain knowledge suggesting that an environment's rewards align with some other point in an interval, the ideas generalize in that discounting should be properly exponentiated to reflect this information.

Regarding avenues for future work, the integral approximation perspective suggests opportunity to explore return modifications corresponding to other integral approximation techniques. If one were to additionally track predecessor rewards, it opens up the possibility of interpolation-based approximations like the trapezoidal rule. Notably, \cite{ayoub2024mitigating} concurrently considered trapezoidal approximations of the Monte Carlo return while exploring \textit{when} to discretize. For the case of exponential discounting, however, we could further leverage that term's closed-form integral.

\subsubsection*{Acknowledgements}
\label{sec:acknowledgements}
This research was generously supported by Amii, NSERC, Google Deepmind, and the Pan-Canadian AI Strategy managed by CIFAR. We would like to thank Forte Shinko, Alan Chan, and Sungsu Lim for insights and discussions contributing to the results in this paper, and the reviewers for valuable feedback during the review process.


\bibliography{main}
\bibliographystyle{apalike}


\newpage

\appendix




\section{Servo Reacher Environment Details}
\label{sec:app_env_details}

The environment state $\mathbf{x}$ is a column vector containing the DC motor's angular velocity [rad/s], the DC motor's current [A], the output shaft's angle [rad], the output shaft's angular velocity [rad/s], and the output shaft's target angle [rad], respectively. The state vector is updated as follows:

\begin{equation*}
    \dot{\mathbf{x}}_t \leftarrow 
\begin{bmatrix}
    -\frac{b_m}{J_m} & \frac{K_t}{J_m} & 0 & 0 & 0 \\
    -\frac{K_t}{L_a} & -\frac{R_a}{L_a} & 0 & 0 & 0 \\
    0 & 0 & 0 & 1 & 0 \\
    -\frac{b_m}{J_m N \eta} & \frac{K_t}{J_m N \eta} & 0 & 0 & 0 \\
    0 & 0 & 0 & 0 & 0
\end{bmatrix} \mathbf{x}_t + 
\begin{bmatrix}
0 \\
\frac{1}{L_a} \\
0 \\
0 \\
0
\end{bmatrix}
A_t
\end{equation*}
\begin{equation*}
    \mathbf{x}_{t+1} \leftarrow \mathbf{x}_t + \dot{\mathbf{x}}_t \Delta_s
\end{equation*}
where $\Delta_s = 10^{-4}$ [s] is the simulation discretization granularity, and $A_t$ is an input voltage with built-in saturation limits of $\in [-12, 12]$ [V]. The output shaft angle is clamped $\in [-1.306, 1.306]$ [rad] in accordance with \cite{mahmood2018benchmarking}. The physical parameters used are detailed below:

\begin{tabular}{|c|c|c|}
    \hline
    $L_a$ & Armature Inductance & $2.05\times10^{-3}$ [H] \\
    $R_a$ & Armature Resistance & 8.29 [Ohm] \\
    $J_m$ & Rotor Inertia & $8.67\times10^{-8}$ [$\textrm{kg}\cdot\textrm{m}^2$] \\
    $b_m$ & Rotor Friction & $8.87\times10^{-8}$ [$\textrm{N}\cdot\textrm{m}\cdot\textrm{s}$] \\
    $K_t$ & Torque Constant & 0.0107 [$\frac{\textrm{N}\cdot\textrm{m}}{\textrm{A}}$] \\
    $N$ & Gear Ratio & 200 \\
    $\eta$ & Gear Efficiency & 0.836 \\
    \hline
\end{tabular}

Given a target discretization interval $> 10^{-4}$ [s], the above updates are repeated until the target elapsed time is reached, keeping track of any overshoot and compensating accordingly in the next time interval. As a reinforcement learning environment, an agent observes the output shaft's angle, angular velocity, and target angle. The initial output shaft angle, $\theta_0$, and target angle, $\theta_{target}$, are uniformly sampled $\in [-1.306, 1.306]$ at the start of each episode, and an episode terminates when $|\theta_{t+1} - \theta_{target}| < 0.1$ [rad] with angular velocity $\dot{\theta}_{t+1} < 0.1$ [rad/s]. An agent provides a continuous-valued action as a voltage, and receives a reward $|\theta_{t+1} - \theta_{target}|$, computed and received jointly with the next observation.

\end{document}